\documentclass[11pt]{article}

\usepackage[preprint]{acl}

\usepackage{amsmath,amsfonts,bm}

\def\eqref#1{equation~\ref{#1}}

\def\1{\bm{1}}

\def\vp{{\bm{p}}}

\def\vx{{\bm{x}}}
\def\vy{{\bm{y}}}
\def\vz{{\bm{z}}}

\DeclareMathAlphabet{\mathsfit}{\encodingdefault}{\sfdefault}{m}{sl}
\SetMathAlphabet{\mathsfit}{bold}{\encodingdefault}{\sfdefault}{bx}{n}

\usepackage{times}
\usepackage{latexsym}
\usepackage[T1]{fontenc}
\usepackage[utf8]{inputenc}
\usepackage{microtype}
\usepackage{inconsolata}
\usepackage{graphicx}
\usepackage{algorithm}
\usepackage{algorithmic}
\usepackage{dsfont}
\usepackage{booktabs}
\usepackage{array}
\usepackage{pifont}
\usepackage{enumitem}
\usepackage{xcolor}

\usepackage[table]{xcolor}
\usepackage{amssymb}
\usepackage{amsthm}
\usepackage{amsmath}
\usepackage{amsfonts}

\theoremstyle{plain}
\newtheorem{proposition}{Proposition}

\newcommand{\ours}{{\fontfamily{ppl}\selectfont ExTra}}

\title{\ours{}: Exploratory Trajectory Optimization for Language Model Reinforcement Learning}

\author{Wenyang Hu\thanks{Correpondence to: wenyang.hu@u.nus.edu}$^{1,2}$, Junxiang Jia$^{2}$, Zhen Shu$^{2}$, Daniel Dahlmeier$^{2}$,\\ \textbf{See-Kiong Ng$^{1}$, Bryan Kian Hsiang Low$^{1}$}\\ 
$^1$National University of Singapore, $^2$SAP\\
}

\begin{document}
\maketitle

\begin{abstract}
Reinforcement Learning with Verifiable Rewards (RLVR) for language-model reasoning can fail at both extremes of task difficulty: easy prompts often produce all-correct, low-diversity rollout groups with little gradient signal, while hard prompts can produce all-incorrect groups with no positive reward. We introduce \textbf{\ours{}} (Exploratory Trajectory Optimization), a GRPO-compatible framework that extracts exploration signals from the model's own rollouts. \ours{} combines two mechanisms: (i)~a novelty reward that adds embedding-based diversity bonuses after GRPO normalization, rewarding diverse \emph{correct} solutions; and (ii)~entropy-guided prefix regeneration, which scores partial trajectories using entropy signals and continues exploration from promising intermediate steps. Across six mathematical reasoning benchmarks, \ours{} improves Qwen3-1.7B over GRPO by about \textbf{+5} points on pass@1 and \textbf{+7} points on pass@16, showing that trajectory-level exploration signals can improve both single-sample accuracy and inference-time coverage. Code is available at:~\href{https://github.com/allen4747/extra}{https://github.com/allen4747/extra}.
\end{abstract}

\section{Introduction}
Reinforcement learning has become a central post-training paradigm for
improving LLM reasoning capabilities~\cite{ouyang2022training,deepseek2025r1}.
Group Relative Policy Optimization (GRPO)~\cite{shao2024deepseekmath} 
is a widely adopted variant that removes the learned value network by
normalizing rewards within a sampled group of trajectories. This design
is effective, but it also exposes a structural weakness at the extremes
of task difficulty, where group-level rewards often become homogeneous.

\paragraph{Two failure modes of GRPO.}When a problem is \emph{easy}, the model generates nearly-uniform correct
responses; the within-group reward variance collapses to near zero,
producing negligible gradients.
The policy then over-optimizes for a narrow set of reasoning patterns,
suppressing trajectory diversity.
This diversity collapse directly degrades pass@$k$ for $k > 1$, a metric that governs inference-time scaling via majority voting or
best-of-$N$ sampling~\cite{brown2024large}.
When a problem is \emph{hard}, the group may contain no correct responses;
all advantages vanish and the policy receives no gradient at all. Together, these failure modes ensure that a substantial fraction of every training batch is either harmful or wasted, directly limiting sample efficiency.

The recent efforts~\cite{yu2025dapo,shrivastava2026sample} (e.g., DAPO) propose mitigations such as difficulty-based data filtering and over-sampling, which (1) discard problems with empirical pass rates outside a target window and (2) keep sampling until the group accuracy is neither 0 nor 1. However, this comes at a steep cost: hard problems, which carry the densest learning signal for frontier reasoning, are discarded entirely, and easy problems, which are retained, still erode diversity over time. More fundamentally, resampling from scratch resets exploration to the beginning of the reasoning chain, ignoring trajectories that the model has already produced, which may already encode useful intermediate structure that informs the solution space.

\paragraph{Exploiting latent signals in trajectories.} This observation points toward a different perspective. Rather than discarding existing trajectories or resampling blindly, we ask: \emph{is there signal already embedded in the model's own generation process that can guide exploration more efficiently?} During every rollout, the model produces two kinds of signal that previous approaches ignored: the \emph{diversity} of previous trajectories, which could distinguish reasoning strategies that are genuinely novel from those that merely paraphrase; and the \emph{structure} of failed trajectories, which encodes how well the intermediate states progressed. We exploit both signals through two complementary mechanisms.




With an embedding-based \emph{novelty reward} that encourages correct solutions that are semantically novel relative to those already seen for the same prompt. We design novelty to be gated on correctness: only correct-and-novel solutions are rewarded, so the policy is never incentivized to produce diverse failures at the expense of task performance. This restores the gradient signal on easy problems while preserving the stability of task reward optimization.

For hard problems, rather than resampling from scratch and exploring the full trajectory space, we novelly propose to regenerate from a promising intermediate reasoning step, effectively continuing exploration from a step already deep in the search space. We identify the most promising reasoning step and use it as a prefix for future generations. Through an embedded signal in the LLM's own generation process, Mean Token Entropy (MTE), the prefix selection requires no supervision and enables guided sampling. It is a validated empirical choice by our experiments in Sec.~\ref{sec:exp:prefix}.


Both mechanisms treat generated trajectories as a source of exploratory signal: novelty rewards optimize for \emph{diversity} across trajectories on easy problems, while entropy-guided regeneration optimizes for \emph{structure} within trajectories on hard problems. Together, they form \textbf{\ours{}} (\textbf{Ex}ploratory \textbf{Tra}jectory Optimization), our novel algorithm that improves LLM reinforcement learning by explicitly optimizing trajectories to be both diverse and structurally promising, operating entirely within the standard GRPO loop without sacrificing sample efficiency, additional reward models, or supervision beyond the binary correctness signal.

We provide theoretical analysis to further support how regeneration helps efficient exploration in RLVR. Our Sec.~\ref{sec:method:theory} elaborates that an informative proxy helps reduce the exploration space and guides exploration. Our empirical analysis validates that there exists such an effective proxy, MTE, which requires no supervision and heavy computation.
    
Extensive evaluation on six mathematical reasoning benchmarks (MATH-500, AMC23, Minerva, OlympiadBench, AIME24, and AIME25) shows that
\ours{} improves Qwen3-1.7B over GRPO by about \textbf{+5} on pass@1 and \textbf{+7} on pass@16. Our ablations and analyses show that the gains come from both improved solution
quality and broader coverage of reasoning trajectories.

\begin{figure*}[ht]
    \centering
    \includegraphics[width=\textwidth]{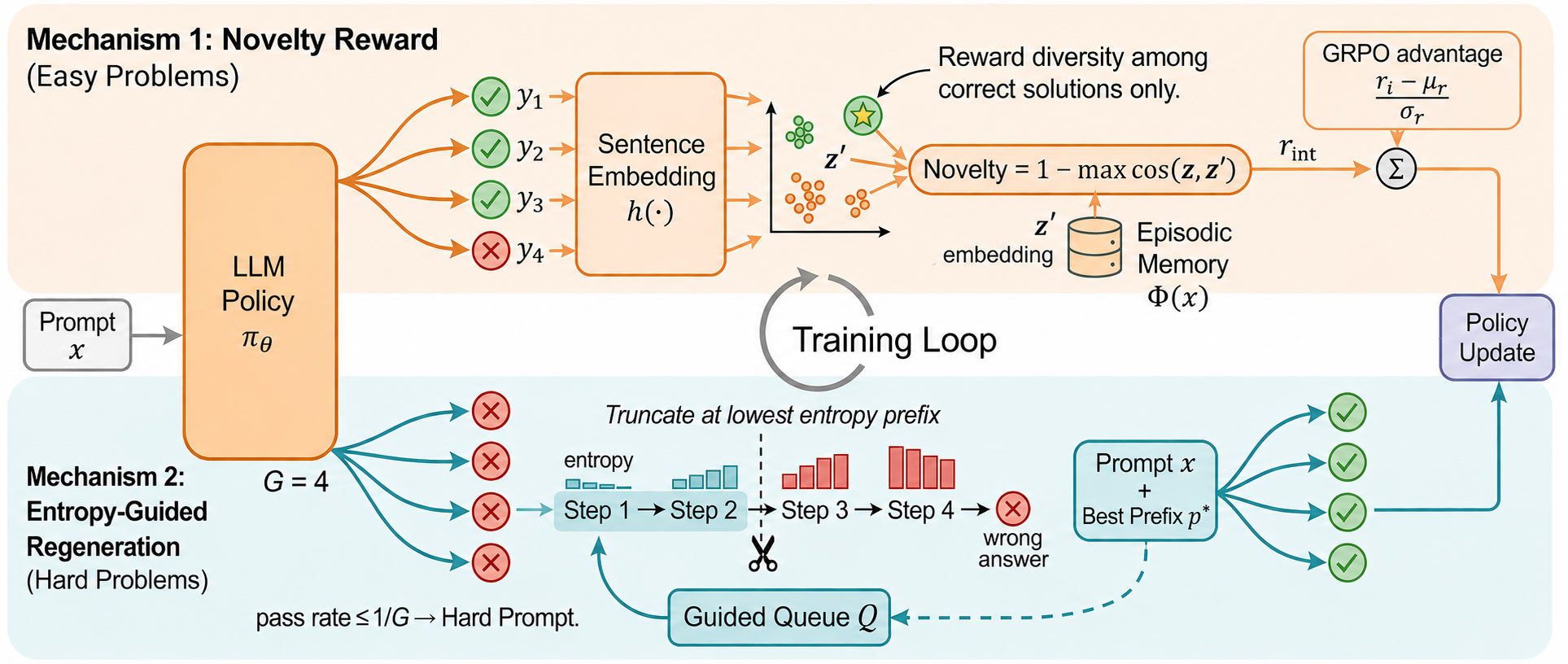}
    \caption{%
      Overview of \ours{}.
      \textbf{Top:} On easy problems, a
      novelty reward is added to GRPO's normalized
      advantage, steering the policy toward diverse correct reasoning
      strategies.
      \textbf{Bottom:} On hard problems, all rollout prefixes are scored by Mean Token Entropy; The lowest-entropy prefix is re-queued as a guided prompt for the next batch, encouraging exploration from a promising intermediate reasoning step.
    }
    \label{fig:overview}
\end{figure*}

\section{Related Work}
\paragraph{RL for LLM Reasoning.}
PPO~\cite{schulman2017proximal} was first applied to align LLMs via
RLHF~\cite{ouyang2022training} and later adapted for chain-of-thought
mathematical reasoning~\cite{wei2022chain}.
GRPO~\cite{shao2024deepseekmath} replaces the value network with
within-group reward normalization; DeepSeek-R1~\cite{deepseek2025r1}
showed that pure GRPO-style RL can elicit strong reasoning from scratch.
DAPO~\cite{yu2025dapo} improves data efficiency by filtering on
empirical pass rates, but discards hard problems entirely.
ExTra retains the full data distribution and instead modifies the
advantage computation and rollout strategy to extract signal from both
difficulty extremes.

\paragraph{Intrinsic Motivation and Exploration Bonuses.}
Intrinsic rewards are a classical tool for encouraging exploration in
sparse-reward environments~\cite{pathak2017curiosity,burda2019exploration}.
ICM~\cite{pathak2017curiosity} rewards prediction error in a learned
feature space; RND~\cite{burda2019exploration} uses the error of a
fixed random network as a novelty proxy.
Adapting these ideas to LLM sequence generation requires a novelty
measure that operates in the high-dimensional space of natural language
reasoning paths. \citep{dai2025cde} proposes using a novelty reward derived from multi-head critics trained on the policy model. In contrast, \ours{} uses cosine distance between sentence embeddings, which requires no additional training.

\paragraph{Process Supervision and Intermediate Rewards.}
Process reward models (PRMs) provide step-level supervision for
reasoning~\cite{lightman2023lets}; Math-Shepherd~\cite{wang2024math}
automates this via Monte Carlo rollouts from each step.
PRMs are powerful but require either human annotation or a separately-trained reward model, adding significant cost.
ExTra's entropy heuristic serves the same role---identifying high-quality
intermediate states---while being computed at zero marginal cost from
generation logits.

\paragraph{Search and Guided Reasoning.}
Tree-of-Thought~\cite{yao2023tree} applies a systematic tree search over
reasoning steps at \emph{inference time}.
ExTra differs in two respects: it operates at \emph{training time},
enriching the replay buffer with trajectories continuing from promising
prefixes; and it uses a single lightweight selection step rather than
a full tree expansion, making it compatible with standard RL pipelines.

\section{Background}

Let $\vx \in \mathcal{X}$ denote a prompt and
$\vy \in \mathcal{Y}$ a response sampled from policy
$\pi_\theta$.  In RLVR, the reward $R(\vx,\vy)$ is usually
binary, indicating whether the final answer can be verified as correct.
The training objective is
\[
  \max_{\theta}\;
  \mathbb{E}_{\vx \sim \mathcal{D},\,
               \vy \sim \pi_{\theta}(\cdot|\vx)}
  \bigl[R(\vx,\vy)\bigr].
\]

Policy-gradient methods optimize this objective through an advantage
$A(\vx,\vy)$, which measures how much better a sampled
response is than the policy's expected response for the same prompt.  The
policy-gradient theorem~\cite{sutton1999policy} gives
\[
\nabla_\theta\, \mathbb{E}\bigl[R(\vx,\vy)\bigr]
\propto
\mathbb{E}\bigl[A(\vx,\vy)\,
\nabla_\theta \log \pi_\theta(\vy|\vx)\bigr].
\]
Thus, positive-advantage responses are reinforced, negative-advantage
responses are suppressed, and near-zero advantages produce little update.

PPO~\cite{schulman2017proximal} typically estimates advantages with a
learned value network.  GRPO~\cite{shao2024deepseekmath} removes this
network by sampling a group of $G$ responses
$\{\vy_i\}_{i=1}^G$ from the old policy
$\pi_{\theta_{\mathrm{old}}}$ for the same prompt $\vx$ and
standardizing their rewards:
\[
  \hat{A}_i = \frac{r_i - \mu_r}{\sigma_r},
\]
where $r_i = R(\vx, \vy_i)$, and $\mu_r$ and $\sigma_r$
are the group mean and standard deviation.  When $\sigma_r=0$, we use the
standard zero-advantage convention.  The policy is then updated with a
clipped surrogate objective regularized by KL divergence to a reference
policy $\pi_{\mathrm{ref}}$; the full expression appears in
Appendix~\ref{sec:app:exp_detail}.


\section{ExTra: Exploratory Trajectory Optimization}

Fig.~\ref{fig:overview} illustrates the full \ours{} framework.
The two mechanisms are modular.  Novelty rewards are applied at every
training step, while prefix regeneration is triggered only for prompts
whose within-group pass rate falls below $1/G$ (quantified as hard).

\subsection{Novelty-Driven Intrinsic Rewards}
\label{sec:method:curiosity}
For each prompt $\vx$, we maintain a rollout memory
$\Phi(\vx)$ containing sentence embeddings of previously generated responses and sibling responses from the same rollout group, computed by a lightweight embedding model
$h(\cdot)$. The intrinsic reward for a new response $\mathbf{y}$ with embedding
$\mathbf{z} = h(\mathbf{y})$ is:
\begin{equation}
  r_{\text{int}}(\mathbf{y}) =
    \mathds{1}_{[R(\mathbf{x},\mathbf{y}) > 0]} \cdot
    \Bigl(1 - \max_{\mathbf{z}' \in \Phi(\mathbf{x})} \cos(\mathbf{z}, \mathbf{z}')\Bigr).
  \label{eq:intrinsic}
\end{equation}
The correctness gate $\mathds{1}_{[R > 0]}$ is essential:
novelty is rewarded \emph{only} among correct solutions.
An incorrect response receives no bonus, even if it differs from all previous responses, so \ours{} does not incentivize diverse failures.

\paragraph{Post-normalization application.}
Where to apply the novelty reward matters. Directly adding
novelty to the raw reward before GRPO normalization couples exploration
to the group mean and variance, so the contribution of the binary task
reward is shrunk or distorted. This is supported by our analysis in Appx~\ref{sec:app:modulation}.  We therefore propose to add novelty after group normalization:
\begin{equation}
  \hat{A}_i^{\mathrm{final}} =
    \hat{A}_i^{\mathrm{GRPO}}
    + \gamma\, r_{\mathrm{int}}^{(i)},
  \label{eq:post_norm}
\end{equation}
where $\hat{A}_i^{\mathrm{GRPO}}=(r_i-\mu_r)/\sigma_r$ when
$\sigma_r>0$ and $0$ otherwise.  The coefficient $\gamma$ has a direct
interpretation: it is the additional normalized advantage assigned to a
maximally novel correct response relative to a minimally novel one.

Our theoretical analysis in App.~\ref{sec:app:coverage} shows that each unit of novelty reward corresponds to discovering a previously uncovered correct mode in $\Phi(\vx)$, justifying treating $r_{\text{int}}$ as a coverage signal rather than a generic diversity penalty.
\subsection{Mean Token Entropy as an Intermediate-State Heuristic}
\label{sec:method:mte}

When a rollout group contains no correct response, neither GRPO nor the correctness-gated novelty reward provides a positive signal. To recover useful information from such hard prompts, we need an unsupervised way to identify partial reasoning trajectories that are more likely to lead to a correct completion.

We define the \textbf{Mean Token Entropy (MTE)} of a partial trajectory
$\vy_{\leq t}$ as
\begin{equation}
  \mathcal{H}_{\mathrm{mean}}(\vy_{\leq t})
  = \frac{1}{t} \sum_{s=1}^{t}
    H\!\bigl(\pi_\theta(\cdot|\vx, \vy_{<s})\bigr),
  \label{eq:mte}
\end{equation}
where $H(\pi) = -\sum_{v \in \mathcal{V}} \pi(v)\log\pi(v)$ is the per-step vocabulary entropy. MTE measures the model's average internal uncertainty while producing a prefix. Our hypothesis is that a coherent reasoning chain, even if incomplete, tends to be generated with lower uncertainty than a drifting or inconsistent chain. MTE is available at no additional forward-pass cost because it is computed from rollout logits.

Before using MTE operationally, we validate it in a controlled Monte Carlo
study. Section~\ref{sec:exp:prefix} shows that MTE
($\rho_{\mathrm{within}}=-0.229$, $p=0.002$) is the strongest
within-problem predictor of prefix value across 600 prefixes from hard problems, outperforming 20 alternative metrics, including several that use the ground-truth answer. This confirms our design choice.

\subsection{Entropy-Guided Prefix Regeneration}
\label{sec:method:regen}


For each hard prompt, we extract discrete intermediate reasoning steps from all $G$ rollouts and compute MTE for every prefix.  Individual prefix scores can be noisy, so we smooth them by semantic similarity:
\begin{equation*}
  \tilde{\mathcal{H}}_j = \sum_k w_{jk}\,\mathcal{H}_k, \;
  w_{jk} = \frac{\exp\!\bigl(\cos(e_j, e_k)/\tau\bigr)}
                {\sum_l \exp\!\bigl(\cos(e_j, e_l)/\tau\bigr)},
  \label{eq:smooth}
\end{equation*}
where $\mathcal{H}_j \equiv \mathcal{H}_{\mathrm{mean}}(\vp_j)$
is the raw MTE of prefix $\vp_j$, $e_j=h(\vp_j)$ is its embedding, and $\tau$ is a smoothing parameter.  The smoothed score lets semantically similar prefixes share information.  We select $\vp^* = \arg\min_j \tilde{\mathcal{H}}_j$, enqueue the guided prompt $[\vx,\vp^*]$, and inject it into a later training
batch.  New suffixes are then sampled as
$\vy_{>t}\sim\pi_\theta(\cdot|\vx,\vp^*)$.  The fixed prefix is
treated as context: verification is applied to the concatenated
trajectory $(\vp^*,\vy_{>t})$, while the policy-gradient loss is
computed only on newly generated suffix tokens.  This focuses exploration on an intermediate state that the model already reached with relatively
low uncertainty, increasing the chance of obtaining a rewarded
continuation.  The full procedure is given in Algorithm~\ref{alg:extra_full}
(Appendix~\ref{app:full_alg}).

\paragraph{Interaction with novelty.}
Regeneration is most useful when the rollout pool contains multiple
plausible reasoning paths.  The novelty reward helps maintain this pool;
entropy-guided regeneration then selects among diverse candidates instead
of repeatedly amplifying a single low-entropy pattern.  The ablations in
Section~\ref{sec:exp:ablation} test this interaction directly.

\begin{table*}[t]
\centering
\footnotesize
\setlength{\tabcolsep}{2.8pt} 
\caption{Qwen3-1.7B mathematical reasoning results.  pass@1 measures
single-sample accuracy; pass@16 measures inference-time coverage across
16 samples.  Best results are bolded and second-best results are
underlined.}
\label{tab:qwen_results}
\begin{tabular}{l cccccc cccccc cc}
\toprule
& \multicolumn{2}{c}{\textbf{AIME24}} & \multicolumn{2}{c}{\textbf{AIME25}} & \multicolumn{2}{c}{\textbf{AMC23}} & \multicolumn{2}{c}{\textbf{MATH-500}} & \multicolumn{2}{c}{\textbf{Minerva}} & \multicolumn{2}{c}{\textbf{Olymp.}} & \multicolumn{2}{c}{\textbf{Avg}} \\
\cmidrule(lr){2-3} \cmidrule(lr){4-5} \cmidrule(lr){6-7} \cmidrule(lr){8-9} \cmidrule(lr){10-11} \cmidrule(lr){12-13} \cmidrule(lr){14-15}
\textbf{Method} & \textbf{@1} & \textbf{@16} & \textbf{@1} & \textbf{@16} & \textbf{@1} & \textbf{@16} & \textbf{@1} & \textbf{@16} & \textbf{@1} & \textbf{@16} & \textbf{@1} & \textbf{@16} & \textbf{@1} & \textbf{@16} \\
\midrule
GRPO & 23.5 & 46.7 & 22.7 & \underline{53.3} & 71.9 & 92.5 & 82.6 & 93.2 & 32.8 & 48.2 & \underline{49.1} & \underline{64.5} & 47.1 & 66.4 \\
DAPO & 26.9 & 53.3 & \underline{29.0} & 43.3 & 72.5 & 92.5 & 76.5 & 88.4 & 29.0 & 46.3 & 43.1 & 57.9 & 46.2 & 63.6 \\
\midrule
ExTra-R & 26.9 & 46.7 & 22.3 & 43.3 & 72.5 & \underline{95.0} & 83.7 & 92.6 & 34.2 & \underline{50.7} & 48.4 & 62.8 & 48.0 & 65.2 \\
ExTra-N & \textbf{31.7} & \underline{63.3} & 27.7 & 50.0 & \underline{74.7} & \textbf{97.5} & \underline{86.0} & \underline{94.8} & \underline{34.8} & \underline{50.7} & \textbf{52.1} & \textbf{67.1} & \underline{51.1} & \underline{70.6} \\
\rowcolor{green!8} \textbf{ExTra (full)} & \underline{31.5} & \textbf{66.7} & \textbf{30.4} & \textbf{63.3} & \textbf{76.4} & \underline{95.0} & \textbf{86.3} & \textbf{95.8} & \textbf{35.2} & \textbf{51.5} & \textbf{52.1} & \textbf{67.1} & \textbf{52.0} & \textbf{73.2} \\
\bottomrule
\end{tabular}
\end{table*}

\subsection{Why Regeneration Helps: A Search-Space Analysis}
\label{sec:method:theory}

Regeneration admits a simple local justification on hard prompts. Let $\mathcal{Y}^*(\vx)\subseteq\mathcal{V}^L$ denote the correct trajectories of length $L$ over vocabulary $\mathcal{V}$. Fixing the first $t$ tokens shrinks the reachable trajectory space from $\mathcal{V}^L$ to $\mathcal{V}^{L-t}$; the gain depends on whether the surviving slice contains correct trajectories at higher density than the original. We compare two sampling strategies for the same policy $\pi_\theta$: \textbf{scratch sampling}, which generates a full trajectory $\vy\sim\pi_\theta(\cdot|\vx)$ with success probability $p_0$; and \textbf{regeneration from prefix $\vp$}, which samples only the suffix $\vy_{>t}\sim\pi_\theta(\cdot|\vx,\vp)$ with conditional success probability $q(\vp)$. The two are linked via the law of total probability: $p_0=\mathbb{E}_{\vp\sim\pi_\theta(\cdot|\vx)}[q(\vp)]$. To show how regeneration helps, we provide a search-space analysis (proof appears in App.~\ref{sec:app:theory}):

\begin{proposition}[Regeneration efficiency]
\label{prop:regen}
With $p_0$ and $q(\cdot)$ as above, $p_0=\mathbb{E}_{\vp\sim\pi_\theta(\cdot|\vx)}[q(\vp)]$, so at least one length-$t$ prefix satisfies $q(\vp)\geq p_0$.  For any selected $\vp^*$ with $q(\vp^*)>p_0$, $K$ independent regenerated attempts yield $P_{\mathrm{regen}}(K)=1-(1-q(\vp^*))^K > 1-(1-p_0)^K=P_{\mathrm{scratch}}(K)$ for every $K\geq 1$, and the expected attempts until the first correct trajectory drop from $1/p_0$ to $1/q(\vp^*)$.
\end{proposition}

\noindent\emph{Implication.}  Scratch sampling marginalizes over all
length-$t$ prefixes weighted by their generation probability;
regeneration replaces this mixture with a single conditional slice.
Because $p_0$ is the prefix-weighted average of $q(\vp)$, an
above-average prefix ($q(\vp)>p_0$) is guaranteed to exist, and
concentrating the budget on it strictly raises the per-attempt success
rate while exploring an exponentially smaller space.  The benefit
therefore reduces to one empirical question: can we identify an
above-average prefix without ground-truth supervision? Sec.~\ref{sec:exp:prefix} answers this in the affirmative for MTE, which beats many alternatives including supervised ones and identifies above-average prefixes. This is exactly the condition required by Prop.~\ref{prop:regen}.

\section{Experiments}
In this section, we evaluate whether \ours{} improves RLVR training along three axes: \textit{(i)} final reasoning accuracy, measured by \textbf{pass@1}; \textit{(ii)} inference-time coverage, measured by \textbf{pass@16}; and \textit{(iii)} exploration efficiency, measured by diversity and rollout cost.

\paragraph{Training setup.}
We train Qwen3-1.7B~\cite{qwen3report} and Nemontron-1.5B on MATH-DAPO~\cite{yu2025dapo}. Unless otherwise stated, all methods use group size $G=6$, batch size $512$, learning rate $3\times10^{-6}$, and train for 250 steps.  Full training details, decoding parameters, and implementation details are given in App.~\ref{sec:app:exp_detail}.

\paragraph{Evaluation.} We evaluate on six mathematical reasoning benchmarks: MATH-500, AMC23, Minerva, OlympiadBench, AIME24, and AIME25.  We report pass@1, the expected accuracy of a single sampled response, and pass@16, the fraction of problems for which at least one of 16 sampled responses is correct. pass@16 is central to our setting because it captures whether additional sampling exposes correct trajectories that pass@1 may miss; the diversity analysis below tests whether these gains correspond to broader trajectory coverage rather than only higher single-sample accuracy.

\paragraph{Baselines and variants.}
We compare against standard GRPO~\cite{shao2024deepseekmath} and DAPO~\cite{yu2025dapo}.  To isolate the two mechanisms in \ours{}, we also evaluate two ablations: \textbf{ExTra-R}, which uses entropy-guided prefix regeneration only, and \textbf{ExTra-N}, which uses the correctness-gated novelty reward only.  \textbf{ExTra} denotes the full method.

\subsection{Main Results}
\label{sec:exp:main}

Tab.~\ref{tab:qwen_results} reports Qwen3-1.7B results for models trained by all five methods across the six benchmarks. It shows \ours{} achieves the highest pass@1 performance, lifting the average pass@1 from 47.1\% to 52.0\% (\textbf{+4.9}) on six benchmarks. The improvement is consistent on every benchmark but heavily skewed toward the hardest benchmarks: \textbf{+8.0} on AIME24 and \textbf{+7.7} on AIME25. 

The pass@16 results show an even clearer exploration benefit.  Because pass@16 measures whether any of 16 sampled solutions is correct, it is sensitive to the breadth of the policy's correct-solution support as well as to single-sample accuracy.  \ours{} improves average pass@16 from 66.4\% to 73.2\% (\textbf{+6.8} points), with particularly large gains on AIME24 (\textbf{+20.0}) and AIME25 (\textbf{+10.0}).  Together with the diversity analysis in Sec.~\ref{sec:exp:diversity}, this suggests that \ours{} does not merely sharpen one preferred reasoning path; it broadens the set of correct trajectories reachable at inference time.

\subsection{Mechanism Decomposition}
\label{sec:exp:mechanism}

The bottom rows of Table~\ref{tab:qwen_results} isolate the contribution of each mechanism.  Relative to GRPO, the regeneration-only variant (\textbf{ExTra-R}) improves average pass@1 by \textbf{+0.9} points but reduces average pass@16 by \textbf{-1.2} points.  This pattern is consistent with the role of regeneration: conditioning on selected prefixes can improve solution quality, but by itself it can also narrow the set of explored trajectories.

\textbf{ExTra-N}, which uses only the correctness-gated novelty reward, achieves consistently higher performance on both average pass@1 (\textbf{+4.0}) and pass@16 (\textbf{+4.2}). This confirms that rewarding correct-but-novel trajectories directly counteracts the diversity collapse induced by homogeneous correct groups.

The full method combines the strengths of both components.  Its average pass@16 gain over GRPO is \textbf{+6.8}, exceeding the sum suggested by the individual deltas ($-1.2 + 4.2 = +3.0$).  This super-additive effect matches the intended interaction: novelty expands the pool of plausible reasoning paths, and regeneration then selects low-entropy prefixes from that richer pool. Without novelty, regeneration can repeatedly amplify similar prefixes; with novelty, it can exploit a broader set of candidate states.

\subsection{Transfer Across Base Models}
\label{sec:exp:model_transfer}

To test whether the gains are specific to Qwen3-1.7B, we repeat the main comparison on Nemotron-1.5B.  Table~\ref{tab:nemontron_results} shows that \ours{} again achieves the best average pass@1 and pass@16.  The absolute gains are smaller than on Qwen3-1.7B, but the direction is consistent: average pass@1 increases from 57.9\% to 58.9\%, and average pass@16 increases from 77.2\% to 79.8\%.  The strongest gains again appear on harder benchmarks, including \textbf{+4.4} pass@1 and \textbf{+6.6} pass@16 on AIME24.

\begin{table*}[t]
\centering
\footnotesize
\setlength{\tabcolsep}{2.8pt} 
\caption{Nemontron-1.5B mathematical reasoning results.  pass@1 measures
single-sample accuracy; pass@16 measures inference-time coverage across
16 samples.  Best results are bolded and second-best results are
underlined.}
\label{tab:nemontron_results}
\begin{tabular}{l cccccc cccccc cc}
\toprule
& \multicolumn{2}{c}{\textbf{AIME24}} & \multicolumn{2}{c}{\textbf{AIME25}} & \multicolumn{2}{c}{\textbf{AMC23}} & \multicolumn{2}{c}{\textbf{MATH-500}} & \multicolumn{2}{c}{\textbf{Minerva}} & \multicolumn{2}{c}{\textbf{Olymp.}} & \multicolumn{2}{c}{\textbf{Avg}} \\
\cmidrule(lr){2-3} \cmidrule(lr){4-5} \cmidrule(lr){6-7} \cmidrule(lr){8-9} \cmidrule(lr){10-11} \cmidrule(lr){12-13} \cmidrule(lr){14-15}
\textbf{Method} & \textbf{@1} & \textbf{@16} & \textbf{@1} & \textbf{@16} & \textbf{@1} & \textbf{@16} & \textbf{@1} & \textbf{@16} & \textbf{@1} & \textbf{@16} & \textbf{@1} & \textbf{@16} & \textbf{@1} & \textbf{@16} \\
\midrule
GRPO & 47.9 & 76.7 & 39.8 & \textbf{70.0} & \textbf{89.2} & 97.5  & 85.9 & 95.6 & 26.1 & 46.0 & 58.3 & 77.7 & 57.9 & 77.2 \\
DAPO & 44.4 & 76.7 & 39.0 & \textbf{70.0} & 87.5 & 97.5  & 86.2 & 96.0 & \textbf{26.9} & 46.3 & 57.9 & 74.0 & 57.0 & 76.8 \\
\midrule
\rowcolor{green!8} \textbf{ExTra} & \textbf{52.3} & \textbf{83.3} & \textbf{43.1} & \textbf{70.0} & 85.0 & \textbf{100} & \textbf{86.5} & \textbf{96.6}  & 26.8 & \textbf{49.3} & \textbf{59.6} & \textbf{79.7} & \textbf{58.9} & \textbf{79.8} \\
\bottomrule
\end{tabular}
\end{table*}

\subsection{Diversity and Training Dynamics}
\label{sec:exp:diversity}

The pass@16 gains suggest that \ours{} improves inference-time coverage. We verify this directly by measuring diversity among trajectories generated for the same problem.  Table~\ref{tab:diversity_comparison} reports three complementary metrics averaged across the six benchmarks: \emph{Distinct-4} ($\uparrow$), the fraction of unique 4-grams across responses; \emph{Self-BLEU} ($\downarrow$), the mean pairwise BLEU-4 between responses~\cite{zhu2018texygen}; and \emph{LogDet} ($\uparrow$), the determinantal point process log-volume $\log\det(I+\alpha K)$, where $K$ is the cosine-similarity kernel of MiniLM-L6 trajectory embeddings.

\ours{} achieves the best Self-BLEU and LogDet, indicating that its responses are less redundant and occupy a broader semantic region.  DAPO has slightly higher Distinct-4 (0.59 vs. 0.58), but this metric mainly captures surface-level lexical variation.  The stronger semantic measures favor \ours{}, matching the pass@16 results and supporting the claim that \ours{} improves trajectory coverage and diversity.

\begin{table}[t]
    \centering
    \small
    \caption{%
      Diversity comparison using different diversity measures averaged across six benchmarks.
    }
    \label{tab:diversity_comparison}
    \begin{tabular}{lccc}
        \toprule
        Method & Distinct-4 ($\uparrow$)  & Self BLEU ($\downarrow$)  & LogDet ($\uparrow$) \\
        \midrule
        GRPO & 0.54 & 82.84 & 4.29 \\
        DAPO & \textbf{0.59} & 81.63 & 4.47\\
        \textbf{ExTra}   & 0.58 & \textbf{80.6}  & \textbf{4.98} \\
        \bottomrule
    \end{tabular}
\end{table}

\begin{figure}[t]
    \centering
    \includegraphics[width=\linewidth]{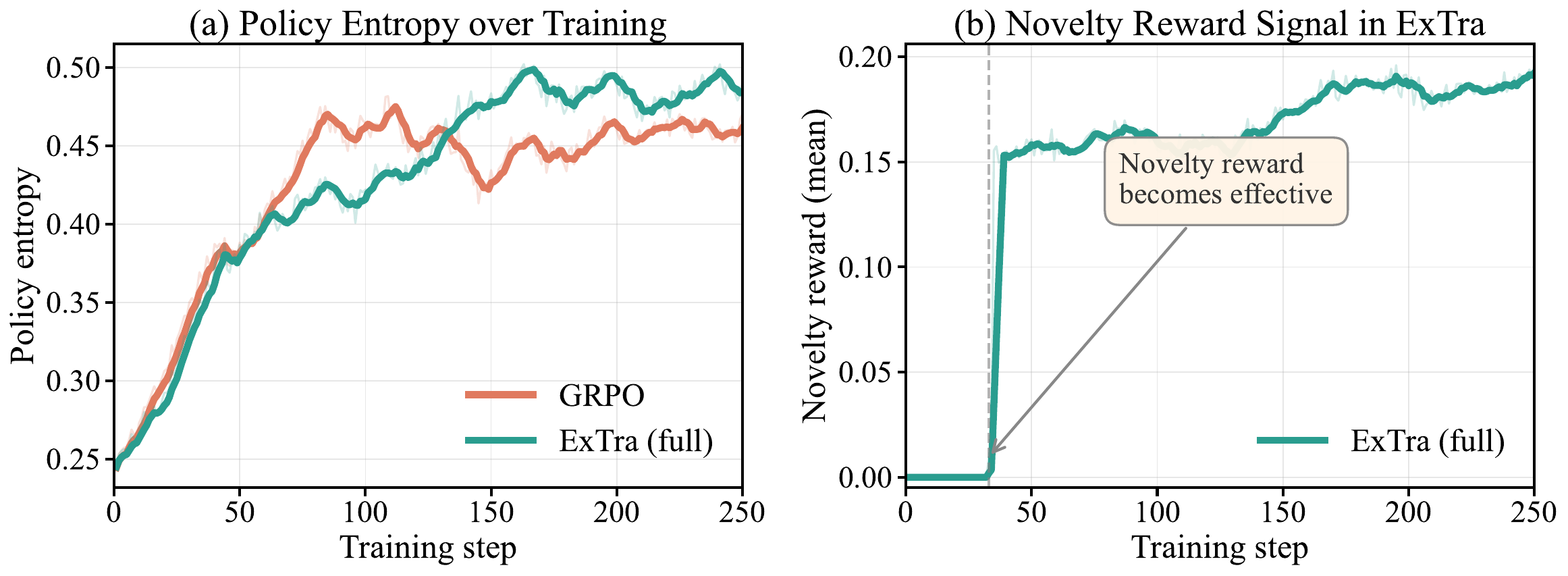}
    \caption{%
      Training dynamics of policy entropy and novelty reward.
      \ours{} sustains an active novelty signal while avoiding the
      entropy collapse associated with homogeneous rollouts.
    }
    \label{fig:training_curves}
\end{figure}

Figure~\ref{fig:training_curves} provides a complementary training-time view for diversity.  \ours{} maintains a non-trivial novelty signal throughout the training steps while avoiding the entropy collapse observed under plain GRPO. The final pass@16 improvements therefore do not appear to be a transient early-training effect; they reflect a sustained change in the policy's exploration behavior.

\subsection{Sample Efficiency}
\label{sec:exp:sample_efficiency}

A central goal of \ours{} is to improve exploration without the large rollout overhead of dynamic resampling.  We therefore compare methods by the \emph{cumulative number of generated prompt instances}, counted before multiplying by the group size $G$.  Each prompt instance produces $G$ responses, and we count it whether or not it contributes to an update. GRPO and \ours{} remain close to one prompt batch per training step, whereas DAPO resamples groups whose rewards collapse to a single value; in our run, DAPO requires an average of $3.05$ generation batches per step.

\begin{figure}[htbp]
  \centering
  \begin{minipage}[t]{0.23\textwidth}
    \centering
    \includegraphics[width=\textwidth]{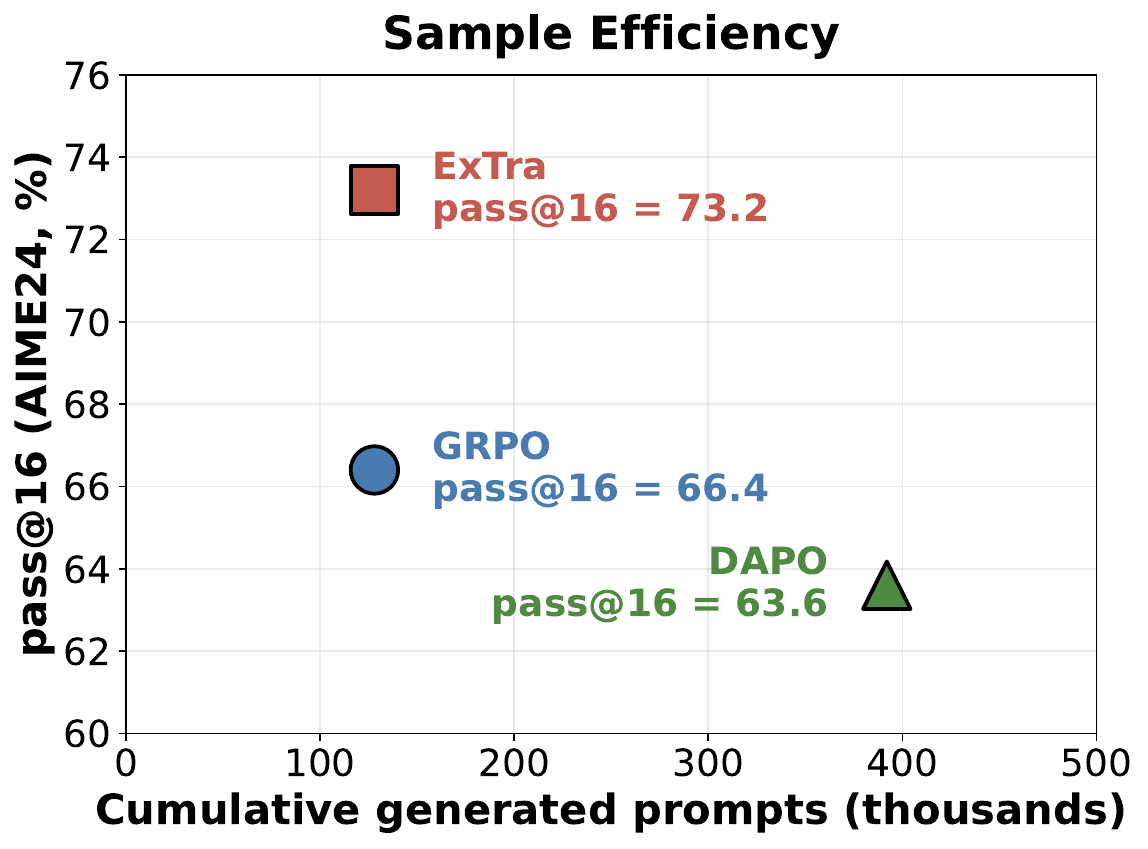} 
    \caption{Sample efficiency comparison at training phase, measured by the cumulative number of prompts sampled.}
    \label{fig:sample-efficiency}
  \end{minipage}
  \hfill
  \begin{minipage}[t]{0.23\textwidth}
    \centering
    \includegraphics[width=\textwidth]{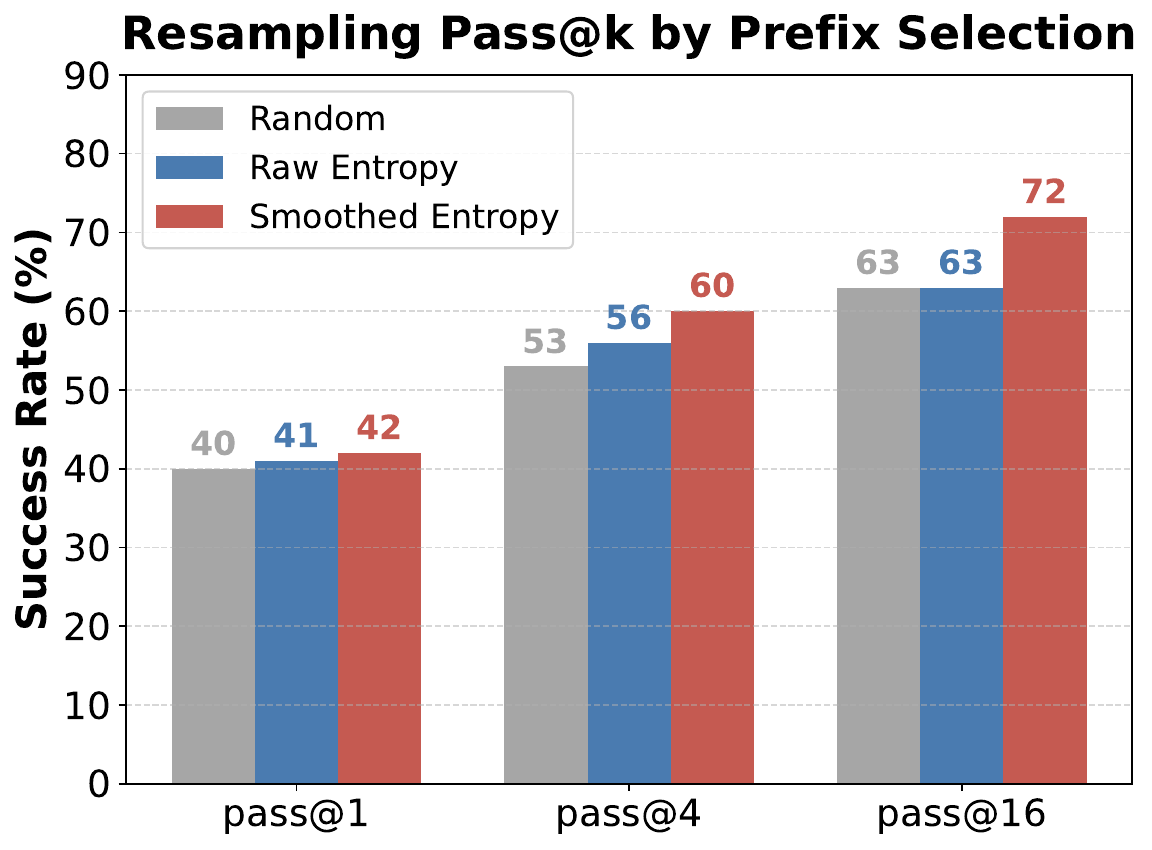}
    \caption{Selecting low-entropy prefixes improves pass@$k$ over random selection, and smoothed entropy further improves it.}
    \label{fig:entropy_analysis}
  \end{minipage}
\end{figure}

Fig.~\ref{fig:sample-efficiency} plots average pass@16 against this prompt budget at step 250.  \ours{} reaches $73.2\%$ average pass@16 using $136$k generated prompt instances.  DAPO uses $392$k prompt instances but reaches only $63.6\%$, while GRPO reaches $66.4\%$ with $128$k prompt instances.  Thus, \ours{} improves over GRPO at a similar rollout cost and outperforms DAPO on both accuracy and sample efficiency.

\subsection{Validating Mean Token Entropy as a Prefix Heuristic}
\label{sec:exp:prefix}

The validity of entropy-guided regeneration hinges on whether MTE reliably identifies prefixes likely to lead to correct completions.  We test this directly in a controlled Monte Carlo study before using the heuristic inside the training loop.

\paragraph{Setup.} We extract 600 prefixes from level 4--5 MATH-500 problems.  For each prefix, we sample $K=16$ continuations and use the empirical pass rate as an estimate of the prefix's Monte Carlo value.  We then evaluate 21 candidate prefix metrics, including unsupervised signals such as entropy, perplexity, and layer consistency, as well as ground-truth-aware signals such as conditional perplexity to the gold answer and semantic similarity to the gold solution.  We rank metrics by Spearman correlation with the Monte Carlo value.  The primary statistic is within-problem correlation, because regeneration selects among prefixes belonging to the same prompt. Full protocol details appear in Appx.~\ref{sec:app:mc_protocol}.

\paragraph{Results.} Table~\ref{tab:prefix_metrics_correlation} reports a representative subset of the metrics.  MTE is the strongest predictor, achieving $\rho_{\mathrm{within}}=-0.229$ ($p=0.002$).  The absolute correlation is moderate, reflecting the inherent noisiness of Monte Carlo prefix evaluation, but its magnitude is more than twice as large as the strongest ground-truth-aware metric in the full appendix table.  This is notable because several of those baselines use the gold answer, whereas MTE is label-free, requires no reward model, and is computed from logits already produced during rollout.

\begin{table}[t]
    \centering
    \small
    \caption{Spearman correlation ($\rho$) between prefix metrics and
    Monte Carlo pass rate over 600 prefixes from 22 problems. Within-problem
    correlation controls for problem-level difficulty and matches the
    setting in which regeneration selects prefixes. Full table in
    App.~\ref{sec:app:full_table}.}
    \label{tab:prefix_metrics_correlation}
    \begin{tabular}{lcc}
        \toprule
        \textbf{Metric} & \textbf{Within $\rho$} & \textbf{Pooled $\rho$} \\
        \midrule
        \multicolumn{3}{l}{\textit{Unsupervised}} \\
        \textbf{Mean Token Entropy} & $\mathbf{-0.229}$ & $\mathbf{-0.141}$ \\
        Layer Consistency & $-0.079$ & $-0.047$ \\
        Unconditional Perplexity & $+0.000$ & $-0.012$ \\
        \midrule
        \multicolumn{3}{l}{\textit{Ground-truth aware}} \\
        Conditional PPL to GT & $-0.058$ & $+0.023$ \\
        GT Log-Probability & $+0.058$ & $-0.023$ \\
        Semantic Similarity to GT & $-0.011$ & $+0.100$ \\
        \bottomrule
    \end{tabular}
\end{table}

These findings support the empirical condition required by Proposition~\ref{prop:regen}: when MTE selects above-average prefixes, regeneration can increase the conditional success probability relative to sampling from scratch.  Figure~\ref{fig:entropy_analysis} confirms this operationally: low-entropy selection improves over random prefix selection, and embedding-smoothed entropy closes much of the gap to oracle selection.

\subsection{Ablation Studies}
\label{sec:exp:ablation}

We conduct two additional ablations to test whether the novelty reward is robust to its coefficient and whether the form of the novelty signal matters.

\paragraph{Sensitivity to the novelty coefficient.} Table~\ref{tab:gamma_sensitivity} sweeps the intrinsic-reward coefficient $\gamma$ while keeping the regeneration mechanism fixed.  Setting $\gamma=0$ recovers ExTra-R.  Moderate novelty pressure improves both pass@1 and pass@16, with the default $\gamma=1.0$ giving the best macro-average result.  A much larger value, $\gamma=5.0$, substantially hurts performance, suggesting that novelty should complement rather than dominate the extrinsic correctness reward.

\begin{table}[t]
    \centering
    \small
    \caption{Sensitivity to the intrinsic-reward coefficient $\gamma$.
    Macro-averaged pass@1 / pass@16 over the six benchmarks.}
    \label{tab:gamma_sensitivity}
    \begin{tabular}{lcc}
        \toprule
        $\gamma$ & \textbf{Avg pass@1} & \textbf{Avg pass@16} \\
        \midrule
        $0.0$ (= \ours{}-R) & 48.0 & 65.2 \\
        $0.5$ & 48.2 & 68.2 \\
        $1.0$ (default) & \textbf{52.0} & \textbf{73.2} \\
        $5.0$ & 41.6 & 60.9 \\
        \bottomrule
    \end{tabular}
\end{table}

\paragraph{Choice of novelty signal.} Table~\ref{tab:novelty_metric} compares three novelty signals on top of entropy-guided regeneration: policy surprise, 4-gram novelty, and the default embedding-based novelty.  Embedding novelty performs best, especially on pass@16.  Compared with 4-gram novelty, it is only +$1.0$ pp better on pass@1 but +$5.0$ pp better on pass@16.  This gap suggests that lexical novelty is not enough: surface-level variation can produce paraphrased solutions that remain close in reasoning space, whereas embedding novelty better captures semantically distinct solution paths.

\begin{table}[t]
    \centering
    \small
    \caption{Novelty-signal ablation. Macro-averaged pass@1 and pass@16.}
    \label{tab:novelty_metric}
    \begin{tabular}{lcc}
        \toprule
        \textbf{Novelty signal} & \textbf{Avg pass@1} & \textbf{Avg pass@16} \\
        \midrule
        Policy surprise & 49.4 & 67.9 \\
        N-gram count & 51.0 & 68.2 \\
        Embedding (default) & \textbf{52.0} & \textbf{73.2} \\
        \bottomrule
    \end{tabular}
\end{table}

\section{Discussion}
\label{sec:discussion}

\textbf{Trajectory-level signals matter.} \ours{} is motivated by a simple observation: GRPO often discards useful trajectory-level information when rewards collapse to all-correct or all-incorrect groups.  Rather than filtering these cases away, \ours{} extracts signal from them.  Correctness-gated novelty separates diverse correct solutions on easy prompts, while entropy-guided regeneration reuses promising partial solutions on hard prompts.  This makes exploration an explicit part of the RL objective rather than only a data selection problem.

\textbf{The design avoids degenerate exploration.} Novelty does not encourage arbitrary behavior because the intrinsic bonus is applied only to correct responses.  Regeneration alone can narrow the search, but the full method mitigates this by pairing regeneration with novelty: novelty broadens the prefix pool, and regeneration focuses sampling within that broader pool.  The ablations support this interaction.

\textbf{\ours{} improves reasoning coverage.} Overall, the results suggest that \ours{} improves not only final-answer accuracy but also the coverage of correct reasoning trajectories.  Its main significance is therefore conceptual as well as empirical: generated rollouts should be treated as sources of exploration signal, not merely as samples to be scored by a binary reward.
\section{Conclusion}

\ours{} is an RLVR training framework that improves exploration in GRPO-style optimization.  It addresses two common failure modes: diversity collapse on easy prompts and gradient starvation on hard prompts.  It combines a correctness-gated novelty reward with entropy-guided prefix regeneration, allowing the model to use signals already present in its own rollouts.  Experiments on mathematical reasoning benchmarks show that \ours{} improves both pass@1 accuracy and pass@$k$ coverage over standard baselines, suggesting that RLVR can benefit from treating generated trajectories as sources of exploration signal.

\bibliography{custom}

\clearpage
\appendix
\onecolumn

\section{Full ExTra Algorithm}
\label{app:full_alg}

Algorithm~\ref{alg:extra_full} gives the full training procedure for
\ours{}.  The algorithm has two additions to standard GRPO: a
post-normalization novelty bonus for correct trajectories and a
prefix-regeneration queue for hard prompts.

\begin{algorithm}[t]
\caption{\ours{}: Exploratory Trajectory Optimization}
\label{alg:extra_full}
\begin{algorithmic}[1]
\REQUIRE Policy $\pi_\theta$, prompt set $\mathcal{D}$, embedding model $h$,
correct-response memory $\mathcal{M}_{\mathrm{r}}$, prefix memory
$\mathcal{M}_{\mathrm{p}}$, guided-prompt queue $\mathcal{Q}$,
novelty weight $\gamma$, smoothing temperature $\tau$, warmup length
$T_{\mathrm{warm}}$
\FOR{training step $t=1,2,\ldots$}
    \STATE Sample prompts $X \sim \mathcal{D}$
    \IF{$\mathcal{Q}$ is not empty}
        \STATE Replace the tail of $X$ with guided prompts dequeued from $\mathcal{Q}$
    \ENDIF

    \STATE Generate $G$ responses per prompt using $\pi_\theta$
    \STATE Compute extrinsic rewards $R_{\mathrm{ext}}$ and token-level entropies

    \STATE Embed all responses with $h$
    \FOR{each prompt $x$ with responses $\{y_i\}_{i=1}^G$}
        \STATE Let $z_i=h(y_i)$
        \STATE Let $\mathcal{C}_i(x)$ be the set of embeddings in $\mathcal{M}_{\mathrm{r}}(x)$ plus correct sibling-response embeddings from the current group, excluding $z_i$
        \STATE Compute correctness-gated novelty using cosine similarity:
        \[
        r_{\mathrm{int}}^{(i)}=
        \mathds{1}_{[R_{\mathrm{ext}}^{(i)}>0]}
        \left(1-
        \max_{z'\in \mathcal{C}_i(x)}
        \cos(z_i,z')
        \right),
        \qquad \max\varnothing=0
        \]
        \STATE Update $\mathcal{M}_{\mathrm{r}}(x)$ with the current correct-response embeddings, keeping the most recent memory entries
    \ENDFOR

    \STATE Compute GRPO advantages $\hat{A}_i^{\mathrm{GRPO}}=(R_{\mathrm{ext}}^{(i)}-\mu_{\mathrm{group}})/\sigma_{\mathrm{group}}$, setting them to zero when $\sigma_{\mathrm{group}}=0$
    \STATE Add novelty after normalization:
    $\hat{A}_i=\hat{A}_i^{\mathrm{GRPO}}+\gamma r_{\mathrm{int}}^{(i)}$
    \STATE Update $\pi_\theta$ with the clipped GRPO objective using $\hat{A}_i$

    \IF{$t > T_{\mathrm{warm}}$}
        \FOR{each prompt $x$ with group pass rate $0$}
            \STATE Extract reasoning prefixes $\{p_j\}$ from the generated responses
            \STATE Compute each prefix's Mean Token Entropy $\mathcal{H}_j$ and embedding $e_j$
            \STATE Store $(p_j,\mathcal{H}_j,e_j)$ in $\mathcal{M}_{\mathrm{p}}(x)$, keeping at most the configured prefix-memory budget
            \STATE Smooth entropy scores over the current and cached prefixes by embedding similarity as in Eq.~\ref{eq:smooth}
            \STATE Select $p^*=\arg\min_j \tilde{\mathcal{H}}_j$
            \STATE Enqueue the guided prompt $[x,p^*]$ into $\mathcal{Q}$
        \ENDFOR
    \ENDIF
\ENDFOR
\RETURN updated policy $\pi_\theta$
\end{algorithmic}
\end{algorithm}

\section{Implementation and Experimental Details}
\label{sec:app:exp_detail}

\subsection{GRPO Objective}

For completeness, we write the GRPO objective used in our implementation.
For each prompt $\mathbf{x}$, a group of $G$ responses
$\{\mathbf{y}_i\}_{i=1}^G$ is sampled from the old policy
$\pi_{\theta_{\mathrm{old}}}$.  The policy is updated by maximizing
\[
  \mathcal{J}_{\mathrm{GRPO}}(\theta)
  = \mathbb{E}
    \left[
      \frac{1}{G} \sum_{i=1}^G
      \min\!\left(
        \rho_i(\theta)\hat{A}_i,
        \mathrm{clip}\bigl(\rho_i(\theta),1-\epsilon,1+\epsilon\bigr)\hat{A}_i
      \right)
      - \beta\,\mathbb{D}_{\mathrm{KL}}(\pi_\theta\|\pi_{\mathrm{ref}})
    \right],
\]
where
$\rho_i(\theta)=\pi_\theta(\mathbf{y}_i|\mathbf{x})/
\pi_{\theta_{\mathrm{old}}}(\mathbf{y}_i|\mathbf{x})$ is the importance
ratio, $\epsilon$ is the clipping parameter, and $\beta$ controls the KL
penalty.

\subsection{Training Configuration}

We follow the configuration described in Section~\ref{sec:exp:main}; full
hyperparameters are listed in Table~\ref{tab:hyperparameters}.  The
novelty bonus is added after GRPO group normalization, as described in
Eq.~\ref{eq:post_norm}.  This preserves the reward normalization used by
GRPO and prevents the novelty score from changing the group mean or
variance.

\subsection{Novelty Memory and Embedding Details}

For each prompt, the correct-response memory stores embeddings of recent
verified-correct responses.  Embeddings are normalized before cosine
similarity is computed.  In our runs, the relevant cosine similarities are
empirically nonnegative; the theoretical analysis states this as an
assumption, so the intrinsic bonus lies in $[0,1]$.  Unless otherwise
stated, the memory keeps the $G=6$ most recent correct responses per prompt, matching
Tab.~\ref{tab:hyperparameters}.

\subsection{Guided Prompts and Loss Masking}

Prefix regeneration constructs a guided prompt by concatenating the
original problem statement with the selected prefix.  The regenerated
sample consists only of the newly generated suffix.  For reward
computation, we verify the full concatenated trajectory consisting of the
fixed prefix and the generated suffix.  For the policy update, the fixed
prefix is treated as context and the GRPO loss is applied only to the
suffix tokens generated by the current policy.  This masking avoids
training the model to imitate cached prefixes while still allowing the
prefix to guide exploration toward useful intermediate states.

\subsection{Prefix Queue and Memory}

Regeneration is disabled during the first $T_{\mathrm{warm}}=30$ steps.
After warmup, only all-incorrect groups are eligible for prefix
regeneration.  For each eligible prompt, we cache extracted prefixes with
their MTE values and embeddings, retaining at most 128 prefixes per
prompt.  Entropy smoothing is performed within the same prompt over the
current prefixes and the cached prefix memory.  Guided prompts are placed
in a FIFO queue with maximum size 4{,}096 and replace the tail of a later
training batch.

\subsection{Evaluation Configuration}

Evaluation uses temperature $0.7$, top-$p=0.9$, and $n=16$ samples per
problem.  We use a maximum generation budget of 31{,}744 tokens per
response to avoid truncating long reasoning chains.  pass@1 is estimated
as the mean single-sample accuracy over the sampled responses.  pass@16
is the fraction of problems for which at least one of the 16 sampled
solutions is correct.  We use the same answer-extraction and verification
pipeline for all methods.

\begin{table}[h]
    \centering
    \small
    \caption{Hyperparameters used in the main experiments.}
    \label{tab:hyperparameters}
    \begin{tabular}{ll}
        \toprule
        \textbf{Hyperparameter} & \textbf{Value} \\
        \midrule
        \multicolumn{2}{l}{\textit{Model and Data}} \\
        Base models & Qwen3-1.7B; Nemotron-1.5B \\
        Training dataset & MATH-DAPO \\
        Evaluation datasets & MATH-500, AMC23, Minerva, OlympiadBench, AIME24, AIME25 \\
        Max prompt length & 2{,}000 tokens \\
        Max response length & 4{,}096 tokens during training \\
        \midrule
        \multicolumn{2}{l}{\textit{GRPO Training}} \\
        Group size $G$ & 6 \\
        Batch size & 512 \\
        Learning rate & $3\times10^{-6}$ \\
        Training steps & 250 \\
        KL penalty $\beta$ & 0.0 \\
        Entropy coefficient & 0.0 \\
        \midrule
        \multicolumn{2}{l}{\textit{Novelty Reward}} \\
        Novelty coefficient $\gamma$ & 1.0 \\
        Correct-response memory size per prompt & 6 \\
        Embedding model & \texttt{all-MiniLM-L6-v2} \\
        \midrule
        \multicolumn{2}{l}{\textit{Prefix Regeneration}} \\
        Warmup steps & 30 \\
        Smoothing temperature $\tau$ & 0.1 \\
        Max prefixes per prompt & 128 \\
        Max queue size & 4{,}096 \\
        \midrule
        \multicolumn{2}{l}{\textit{Evaluation}} \\
        Sampling temperature & 0.7 \\
        Top-$p$ & 0.9 \\
        Samples per problem & 16 \\
        Max generation tokens & 31{,}744 \\
        Reported metrics & pass@1, pass@16 \\
        \midrule
        \multicolumn{2}{l}{\textit{Infrastructure}} \\
        GPUs & NVIDIA H100 \\
        \bottomrule
    \end{tabular}
\end{table}

\section{Prefix-Value Evaluation Details}
\label{sec:app:prefix_eval}

\subsection{Monte Carlo Prefix Evaluation Protocol}
\label{sec:app:mc_protocol}

Section~\ref{sec:exp:prefix} evaluates whether Mean Token Entropy (MTE)
can identify promising intermediate reasoning states.  The experiment
estimates the value of a prefix by sampling continuations from it and
measuring the resulting pass rate.

\begin{itemize}[leftmargin=*]
  \item \textbf{Problem selection:} 22 MATH-500 problems at difficulty
  levels 4--5 where Qwen3-1.7B produces at least one correct and one
  incorrect full trajectory.
  \item \textbf{Initial trajectories:} 32 full trajectories per problem,
  sampled at temperature 1.0.
  \item \textbf{Prefix extraction:} Each trajectory is truncated at
  20\%, 40\%, 60\%, and 80\% of its reasoning-step count.
  \item \textbf{Continuations:} For each prefix, we sample $K=16$
  continuations using temperature $0.7$ and top-$p=0.9$.
  \item \textbf{Total data:} After parsing, filtering, and per-problem balancing, 600 evaluated prefix--value pairs, corresponding to 9{,}600 sampled continuations in total.
  \item \textbf{Correlation:} We report within-problem Spearman
  correlation, averaged using Fisher $z$-transformation, and pooled
  Spearman correlation over all 600 prefixes.
\end{itemize}

Within-problem correlation is the primary statistic because regeneration
selects among prefixes belonging to the same prompt.  Pooled correlation
is reported only as a secondary diagnostic because it is confounded by
problem-level difficulty.

\subsection{Full Prefix Metric Correlation Table}
\label{sec:app:full_table}

Table~\ref{tab:full_prefix_metrics} reports all 21 evaluated prefix
metrics.  Metrics are grouped by whether they require access to the
ground-truth answer.  MTE is the strongest within-problem predictor even
though it is unsupervised and computed from rollout logits.

\begin{table}[h]
    \centering
    \small
    \caption{Spearman correlation between prefix metrics and Monte Carlo
    estimated pass rate over 600 prefixes from 22 problems.}
    \label{tab:full_prefix_metrics}
    \begin{tabular}{lrr}
        \toprule
        \textbf{Metric} & \textbf{Within $\rho$} & \textbf{Pooled $\rho$} \\
        \midrule
        \multicolumn{3}{l}{\textit{Unsupervised Metrics}} \\
        \textbf{Prefix Token Entropy (Mean)} & $\mathbf{-0.229}$ & $\mathbf{-0.141}$ \\
        Prefix Token Entropy (Std) & $-0.143$ & $-0.140$ \\
        Layer Consistency (Mid vs. Last) & $-0.079$ & $-0.047$ \\
        Logit Lens Answer Prob & $-0.049$ & $-0.055$ \\
        Unconditional Perplexity & $<0.001$ & $-0.012$ \\
        \midrule
        \multicolumn{3}{l}{\textit{Ground-Truth-Aware Metrics}} \\
        Self-Consistency PPL Gap & $+0.098$ & $-0.086$ \\
        GT Log-Probability (Min) & $-0.083$ & $+0.036$ \\
        Mutual Information (Prefix; GT) & $+0.073$ & $-0.004$ \\
        GT Entropy Reduction & $+0.073$ & $-0.004$ \\
        Top-$K$ GT Token Overlap & $-0.065$ & $+0.074$ \\
        GT Tokens in Top-10 Fraction & $+0.061$ & $-0.067$ \\
        GT Log-Probability (Mean) & $+0.058$ & $-0.023$ \\
        GT Answer Prob (Geomean) & $+0.058$ & $-0.023$ \\
        PMI (Prefix, GT) & $+0.058$ & $+0.045$ \\
        Normalized GT Log-Probability & $-0.058$ & $-0.013$ \\
        GT Log-Probability (Sum) & $+0.058$ & $+0.167$ \\
        Relative Perplexity & $+0.058$ & $-0.029$ \\
        Conditional PPL to GT & $-0.058$ & $+0.023$ \\
        Prefix-GT Hidden State Alignment & $+0.044$ & $-0.007$ \\
        PPL to Own Projected Answer & $+0.022$ & $-0.034$ \\
        Semantic Similarity to GT & $-0.011$ & $+0.100$ \\
        \bottomrule
    \end{tabular}
\end{table}

\section{Additional Analysis}
\label{sec:app:additional_analysis}

\subsection{Post-Normalization Novelty Application}
\label{sec:app:modulation}

A key design choice in \ours{} is to add the novelty reward after GRPO's
group-wise normalization rather than adding it to the raw reward before
normalization.  Throughout this section, we assume the empirical regime observed in our
experiments: cosine similarities used in Eq.~\ref{eq:intrinsic} lie in
$[0,1]$, and therefore the correctness-gated novelty bonus $b_i$ lies in
$[0,1]$.  Let $R_i\in\{0,1\}$ be the verifiable reward.  If novelty is added
before normalization, the advantage becomes
\[
\hat A_i^{\mathrm{pre}}
=
\frac{(R_i+\gamma b_i)-\frac{1}{G}\sum_j(R_j+\gamma b_j)}
{\operatorname{Std}_j(R_j+\gamma b_j)}.
\]
This couples novelty to the normalization statistics.  A shared novelty
offset is removed by mean subtraction, while variation in $b_i$ changes
the denominator and can shrink or distort the relative contribution of the
binary task reward.

Post-normalization application avoids this coupling:
\[
\hat A_i^{\mathrm{post}}
=
\frac{R_i-\mu_R}{\sigma_R}+
\gamma b_i,
\]
with the zero-variance convention used for the GRPO term.  The standard
reward normalization is preserved, and $\gamma$ has a stable
interpretation: it is the additional advantage, in normalized units,
assigned to a maximally novel correct solution relative to a minimally
novel one.  Since novelty does not enter the group mean or variance, it
supplements the task reward instead of changing how the task reward is
standardized.

\subsection{Computational Overhead}
\label{sec:app:overhead}

\ours{} adds computation in two places: prefix extraction for
regeneration and response embedding for semantic novelty.  Table~\ref{tab:wallclock}
reports wall-clock overhead per step over 100 training steps.

\begin{table}[h]
    \centering
    \small
    \setlength{\tabcolsep}{6pt}
    \caption{Per-step wall-clock overhead relative to GRPO, reported as mean $\pm$ standard deviation over 100 training
    steps. The exploration phase is unique to \ours{} and includes prefix
    memory updates and novelty computation.}
    \label{tab:wallclock}
    \begin{tabular}{lr}
        \toprule
        \textbf{Method} & \textbf{Additional overhead} \\
        \midrule
        GRPO & 0 s \\
        +Regeneration & $126.9 \pm 10.4$ s \\
        +Regeneration+Embedding novelty & $327.8 \pm 16.1$ s \\
        +Regeneration+N-gram novelty & $164.1 \pm 21.1$ s \\
        \bottomrule
    \end{tabular}
\end{table}

Embedding novelty gives the best pass@16 performance, but it is also the
most expensive option.  N-gram novelty roughly halves the overhead, but
as shown in Table~\ref{tab:novelty_metric}, it loses substantial
inference-time coverage.  We therefore use embedding novelty as the
default and treat lexical novelty as a lower-cost alternative when
throughput is the primary constraint.

\subsection{Scope and Composition with Other Methods}
\label{sec:app:scope}

Our experiments focus on mathematical reasoning with 1--2B-scale models.
The entropy heuristic is likely to transfer best to domains where correct
intermediate reasoning reduces token-level uncertainty.  Code generation
may share this property, while factual QA may be more challenging because
fluent but incorrect chains can still have low entropy.

\ours{} is orthogonal to batch-filtering methods such as DAPO.  DAPO
changes which groups contribute to the gradient update; \ours{} changes
how trajectories are rewarded and regenerated within the batch.  A
natural extension is to combine the two: filtering can remove low-value
updates, while \ours{} can broaden the solution distribution inside the
retained groups.

\section{Theoretical Details}
\label{sec:app:theory}

This section gives local arguments supporting the design of \ours{}.  The
claims are intentionally modest: they do not prove convergence of LLM RL,
but they isolate why the proposed mechanisms are useful under binary
rewards and group-relative normalization.

\subsection{Homogeneous GRPO Degeneracy}
\label{sec:app:homogeneous_grpo}

\paragraph{Lemma A.1.}
Fix a prompt $\vx$ and a GRPO group $\{\vy_i\}_{i=1}^G$ with scalar
rewards $r_i=c$ for every $i$.  Let
\[
\mathcal{L}_{\mathrm{grp}}(\theta)
=
\frac{1}{G}\sum_{i=1}^G
\min\!\left(
\rho_i(\theta)\hat A_i,
\mathrm{clip}(\rho_i(\theta),1-\epsilon,1+\epsilon)\hat A_i
\right)
\]
be the reward-driven part of the clipped GRPO surrogate.  Standard GRPO
sets $\hat A_i=0$ when $\sigma_r=0$ to avoid division by zero; we adopt
this zero-variance convention.  Under it, $\hat A_i=0$ for all $i$, and
therefore $\mathcal{L}_{\mathrm{grp}}(\theta)=0$ and
$\nabla_\theta\mathcal{L}_{\mathrm{grp}}(\theta)=0$ for every $\theta$.
If a KL penalty is used, only that separate regularizer remains.

\paragraph{Proof.}
Since all rewards equal $c$, their group mean is $\mu_r=c$ and every
centered reward is zero.  The zero-variance convention sets all
standardized advantages to zero.  Substituting $\hat A_i=0$ into the
surrogate, both $\rho_i(\theta)\hat A_i=0$ and
$\mathrm{clip}(\rho_i(\theta),1-\epsilon,1+\epsilon)\hat A_i=0$, so
every minimum evaluates to zero.  Hence the reward-driven surrogate is
constant in $\theta$ and has zero gradient.
\hfill$\square$

\subsection{Sign Correctness of Post-Normalization Novelty}
\label{sec:app:sign_correctness}

With binary rewards $R_i\in\{0,1\}$, novelty bonus $b_i\in[0,1]$, and
correctness gate $b_i=0$ whenever $R_i=0$, \ours{} uses
\[
\hat A_i^{\mathrm{final}}=\hat A_i^{\mathrm{GRPO}}+\gamma b_i,
\qquad \gamma\geq 0.
\]
This construction has three useful properties.  First, in any mixed group,
standardization preserves the strict ordering between correct and
incorrect responses, and the correctness gate ensures that incorrect
responses receive no novelty bonus.  Therefore a correct response cannot
be ranked below an incorrect one because of novelty.  Second, in an
all-correct group, the GRPO advantage is zero by Lemma~A.1, so the
novelty term becomes the only source of within-group advantage.  Third,
in an all-incorrect group, both the GRPO advantage and the gated novelty
bonus are zero, so \ours{} does not reward diverse failures.

This is a sanity check on the construction rather than a formal diversity
guarantee; the diversity effect is evaluated empirically in
Sec.~\ref{sec:exp:diversity} and Sec.~\ref{sec:exp:ablation}.

\subsection{Novelty as Correct-Mode Coverage}
\label{sec:app:coverage}

\paragraph{Proposition A.2.}\label{prop:novel}
Fix a prompt $\vx$.  Suppose correct trajectories partition into latent
semantic modes $\mathcal{C}_1,\ldots,\mathcal{C}_M$, and suppose
$\Phi(\vx)$ contains one representative embedding for each already
discovered correct mode and no representatives of undiscovered modes.
Assume ideal separation under cosine similarity: two correct trajectories
in the same mode have similarity $1$, while trajectories in different modes
have similarity $0$.  Let
\[
C(\Phi)=
\left|\{m:\Phi(\vx)\text{ contains a representative from }\mathcal{C}_m\}\right|
\]
be the number of covered correct modes, and let $\Phi^+$ be the memory
after adding a newly sampled trajectory $\vy$ if it is correct.  Here
$r_{\mathrm{int}}^{\mathrm{raw}}$ denotes the \emph{pre-normalization}
intrinsic reward of Eq.~\ref{eq:intrinsic}, distinct from the
post-normalization advantage contribution $\gamma\,r_{\mathrm{int}}^{(i)}$
used in the algorithm.  With $\max\varnothing=0$,
\[
r_{\mathrm{int}}^{\mathrm{raw}}(\vx,\vy;\Phi)
=
\mathds{1}_{[R(\vx,\vy)>0]}
\left(1-\max_{\vz'\in\Phi(\vx)}\cos(h(\vy),\vz')\right)
=
C(\Phi^+)-C(\Phi).
\]

\paragraph{Proof.}
If $\vy$ is incorrect, the correctness gate makes the intrinsic reward
zero and the number of covered correct modes does not change.  If $\vy$
is correct and belongs to an already covered mode, the maximum cosine similarity
to the stored representative of that mode is $1$, so the intrinsic reward
is zero and coverage does not increase.  If $\vy$ is correct and belongs
to an uncovered mode, all stored representatives are from different modes
and have cosine similarity $0$; the intrinsic reward is $1$, and adding $\vy$
increases the coverage count by exactly one. \hfill$\square$

\subsection{Regeneration Efficiency}
\label{sec:app:regen_proof}

\paragraph{Proof of Proposition~\ref{prop:regen} (restated).}
We first restate the claim for a fixed policy: with $p_0$ the
scratch-sampling success probability and $q(\vp)$ the conditional success
probability of the suffix given prefix $\vp$, at least one length-$t$
prefix in the policy support satisfies $q(\vp)\ge p_0$; for any $\vp^*$
with $q(\vp^*)>p_0$,
$P_{\mathrm{regen}}(K)>P_{\mathrm{scratch}}(K)$ for every $K\ge1$, and the
expected number of attempts to obtain a correct trajectory falls from
$1/p_0$ to $1/q(\vp^*)$.

By the law of total probability, conditioning on the length-$t$ prefix of
a scratch rollout gives
\[
p_0
=\sum_{\vp\in\mathcal{V}^t}
\Pr_{\pi_\theta}[\vy_{\leq t}=\vp\mid\vx]q(\vp)
=\mathbb{E}_{\vp\sim\pi_\theta(\cdot|\vx)}[q(\vp)].
\]
Thus $p_0$ is the prefix-weighted mean of $q(\cdot)$.  Therefore at least
one prefix satisfies $q(\vp)\geq p_0$, with strict inequality unless
$q(\cdot)$ is constant over the support.

Now fix a selected prefix $\vp^*$ with $q(\vp^*)>p_0$.  For any strategy
with per-attempt success probability $p$, the probability that all $K$
independent attempts fail is $(1-p)^K$, so the probability of at least
one success is $1-(1-p)^K$.  Substituting $p=p_0$ and $p=q(\vp^*)$ gives
\[
P_{\mathrm{regen}}(K)=1-(1-q(\vp^*))^K
>
1-(1-p_0)^K=P_{\mathrm{scratch}}(K)
\]
for every $K\geq1$.  The expected waiting time until the first success in
an independent Bernoulli sequence with success probability $p>0$ is
$1/p$, so regeneration reduces the expected number of attempts from
$1/p_0$ to $1/q(\vp^*)$.  Finally, fixing the first $t$ tokens reduces
the reachable trajectory space from $\mathcal{V}^L$ to
$\mathcal{V}^{L-t}$ because only the suffix remains to be sampled.  This last statement is a combinatorial search-space comparison; the probability of success is governed by the conditional success rate $q(\vp^*)$. \hfill$\square$

\end{document}